\documentclass[sigconf, review]{acmart}
\usepackage{graphicx}
\usepackage{textcomp}
\usepackage{xcolor}
\usepackage{siunitx}
\usepackage{url}
\usepackage{xcolor}         
\usepackage{amsmath}
\usepackage{color}
\usepackage{makecell}
\usepackage{amsmath}
\usepackage{multirow} 
\usepackage{pifont}
\usepackage{comment}
\usepackage{makecell}
\usepackage{subcaption}
\usepackage[normalem]{ulem}
\usepackage{lipsum}
\usepackage{multirow}
\usepackage{wrapfig}
\usepackage{pifont}
\usepackage{algorithm}
\usepackage{algpseudocode}
\usepackage{booktabs}
\usepackage{amsmath}
\usepackage[table]{xcolor}
\usepackage{pifont}
\newcommand{\xmark}{\ding{55}} 
\AtBeginDocument{%
  }

\setcopyright{acmlicensed}
\copyrightyear{2018}
\acmYear{2018}
\acmDOI{XXXXXXX.XXXXXXX}
\acmConference[Conference acronym 'XX]{Make sure to enter the correct
  conference title from your rights confirmation email}{June 03--05,
  2018}{Woodstock, NY}

\acmSubmissionID{471}

\begin{document}

\title{MorphOPC: Advancing Mask Optimization with Multi-scale Hierarchical Morphological Learning}

\author{Yuting Hu}
\affiliation{%
  \institution{University at Buffalo}
  \city{Buffalo}
  \state{NY}
  \country{USA}
}
\author{Lei Zhuang}
\affiliation{%
  \institution{IBM T. J. Watson Research Center}
  \city{Yorktown Heights}
  \state{NY}
  \country{USA}
}
\author{Chen Wang}
\affiliation{%
  \institution{University at Buffalo}
  \city{Buffalo}
  \state{NY}
  \country{USA}
}
\author{Ruiyang Qin}
\affiliation{%
  \institution{Villanova University}
  \city{Villanova}
  \state{PA}
  \country{USA}
}
\author{Hua Xiang}
\affiliation{%
  \institution{IBM T. J. Watson Research Center}
  \city{Yorktown Heights}
  \state{NY}
  \country{USA}
}
\author{Gi-joon Nam}
\affiliation{%
  \institution{IBM T. J. Watson Research Center}
  \city{Yorktown Heights}
  \state{NY}
  \country{USA}
}
\author{Jinjun Xiong}
\affiliation{%
  \institution{University at Buffalo}
  \city{Buffalo}
  \state{NY}
  \country{USA}
}









\begin{abstract}
As feature sizes shrink to the nanometer scale, accurately transferring circuit patterns from photomasks to silicon wafers becomes increasingly challenging. Optical proximity correction (OPC) is widely used to ensure pattern fidelity and manufacturability. Recent generative mask optimization models based on encoder-decoder architecture can synthesize near-optimal masks, serving as fast machine learning (ML) surrogates for traditional OPC. However, these models often fail to capture the geometric transformations from target layouts to mask patterns, leading to suboptimal quality. In this work, we formulate mask generation as a sequence of morphological operations on local layout features and propose \textit{MorphOPC}, a multi-scale hierarchical model with neural morphological modules to learn these transformations. Experiments on edge-based OPC and ILT benchmarks across metal and via layers show that \textit{MorphOPC} consistently outperforms state-of-the-art methods, achieving higher printing fidelity and lower manufacturing cost, demonstrating strong potential for scalable mask optimization.
\end{abstract}

\settopmatter{printacmref=false}
\setcopyright{none}
\renewcommand\footnotetextcopyrightpermission[1]{}
\pagestyle{plain}

\maketitle

\section{Introduction}
In optical lithography, photomask serves as the blueprint for transferring circuit patterns onto silicon wafers through a sequence of exposure, development, and etching processes \cite{chiu1997optical}. As feature sizes shrink below the exposure wavelength, optical diffraction and process variations lead to significant pattern distortions on the printed wafer \cite{yang2025advancements}. To ensure faithful printing, the mask optimization process modifies the mask geometries such that, the resulting wafer patterns faithfully reproduce the intended target. Optical Proximity Correction (OPC) is the dominant mask optimization technique for compensating lithographic distortions \cite{awad2014fast, kuang2015robust, starikov2003design}. Among OPC techniques, model-based OPC requires numerous iterations of simulation and correction using calibrated optical and resist models, resulting in extremely high computational cost. Although Inverse Lithography Techniques (ILT) \cite{poonawala2007mask} formulate mask synthesis as a continuous optimization problem and can produce more accurate masks, they are still computationally prohibitive for full-chip scale applications due to repeated forward–backward simulations and large memory requirements \cite{jia2010machine}.
\begin{figure}[tbp]
\centerline{\includegraphics[width=\columnwidth]{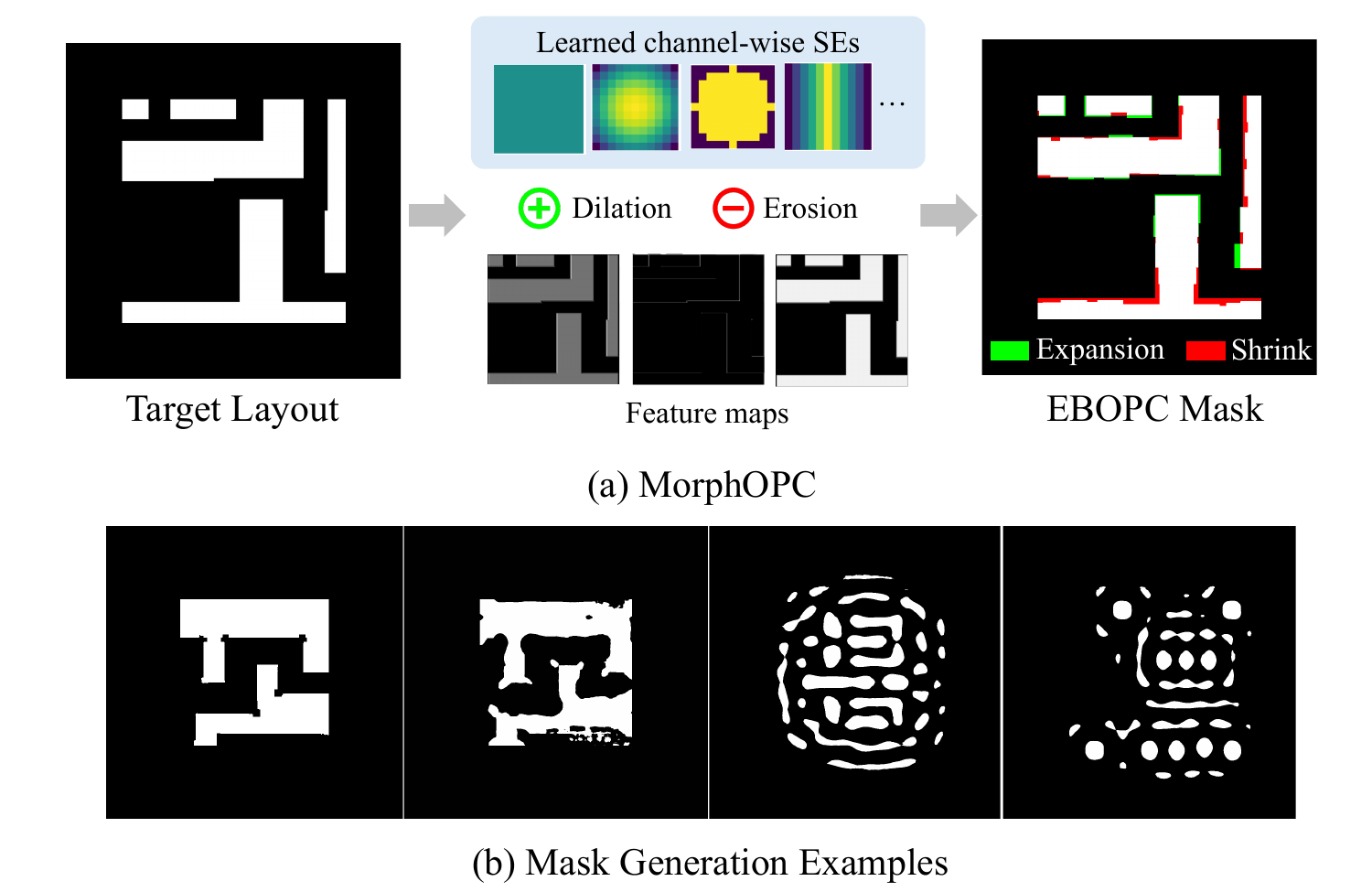}}
\caption{(a) MorphOPC learns morphological operations on local features for mask generation.
(b) MorphOPC supports both pixel-based (ILT) and edge-based (EBOPC) OPC across metal and via layers.}
\label{fig1}
\vspace{-6mm}
\end{figure}

Recently, machine learning (ML) OPC solutions provide significant speedups. GAN-OPC \cite{yang2018gan} employs generative adversarial networks (GAN) with ILT-guided pre-training to directly learn the target-to-mask mapping. DAMO \cite{chen2020damo} combines a high-resolution conditional GAN and a feed-forward correction network with back-propagated lithography gradients to directly generate optimized masks. RL-OPC \cite{10233698} formulates mask optimization as a reinforcement learning problem, where an interactive agent adjusts mask edges on a canvas based on mask quality rewards. Neural-ILT \cite{jiang2020neural} reformulates ILT into a differentiable neural network that jointly optimizes printability and shot count, simplifying mask patterns and reducing cost. CFNO \cite{yang2022large} incorporates lithography physics into a Fourier neural operator for more accurate and data-efficient learning, and EMOGen \cite{zheng2024emogen} enables the co-evolution of pattern generation and ILT models to enhance mask optimization through layout pattern generation. In parallel, optimization-based ILT methods continue to advance mask synthesis through improved formulations and numerical efficiency. MultiILT \cite{sun2023efficient} adopts a multi-level optimization strategy to progressively refine mask patterns from coarse to fine scales, improving convergence and scalability. CurvyILT \cite{yang2025gpu} introduces curvilinear mask representations to better approximate ideal continuous shapes, achieving higher fidelity with fewer shots. More recently, DiffOPC \cite{chen2024differentiable} leverages differentiable lithography modeling and gradient-based optimization to enable end-to-end mask optimization with improved convergence behavior. Despite their strong performance, these methods remain computationally intensive due to repeated lithography simulations and iterative optimization loops.

While generative models excel at learning the mapping from layout patterns to mask shapes, their understanding of geometry is inherently statistical, not analytical. In practice, accurate mask optimization requires explicit geometric reasoning. For example, tip-to-tip patterns require hammerhead extensions to counteract line-end shortening; tip-to-side patterns can induce localized notch effects that must be compensated; and L-shaped corners exhibit distinct printability, inner corners tend to print larger and necessitate anti-serif corrections, whereas outer corners experience rounding and typically require serif extensions to maintain sharpness \cite{mack2008fundamental, gupta2004merits, pawlowski2007fast}. These shape-dependent corrections demand explicit modeling of local geometric context and its associated transformations. To address these limitations, we propose a morphological abstraction of OPC that views mask generation as a composition of morphological operations applied to the target pattern. As illustrated in Figure~\ref{fig1}, morphological operations are non-linear transformations that probe a pattern with a small, predefined shape called a structuring element (SE) to modify its features \cite{haralick1987image,najman2013mathematical,shih2017image}. Specifically, dilation expands the boundaries of a pattern, while erosion shrinks them. These two fundamental operations naturally capture the expansion and contraction behaviors in practical OPC corrections, thereby providing an interpretable framework for modeling shape-dependent target-to-mask transformations. Based on the abstraction, we propose \textit{MorphOPC} to jointly learn both the coefficients for morphological operations and their corresponding SEs directly from the convolutional feature maps. As the corrections are explicitly tied to the fundamental expansion and contraction principles of morphology, our approach ensures that the mask optimization is not only computationally efficient but also explainable.
\begin{itemize}
    \item We formulate mask generation as a composition of learned morphological operations, specifically dilation and erosion, applied to local layout patterns.
    \item We propose \textit{MorphOPC}, a novel multi-scale hierarchical architecture that integrates differentiable morphological modules into an encoder-decoder framework.
    \item Extensive experiments demonstrate that \textit{MorphOPC} achieves SOTA performance across multiple benchmarks and exhibits superior generalization to unseen, complex design patterns compared to existing learning-based OPCs.
\end{itemize}
\begin{figure*}[tbp]
\centerline{\includegraphics[width=\textwidth]{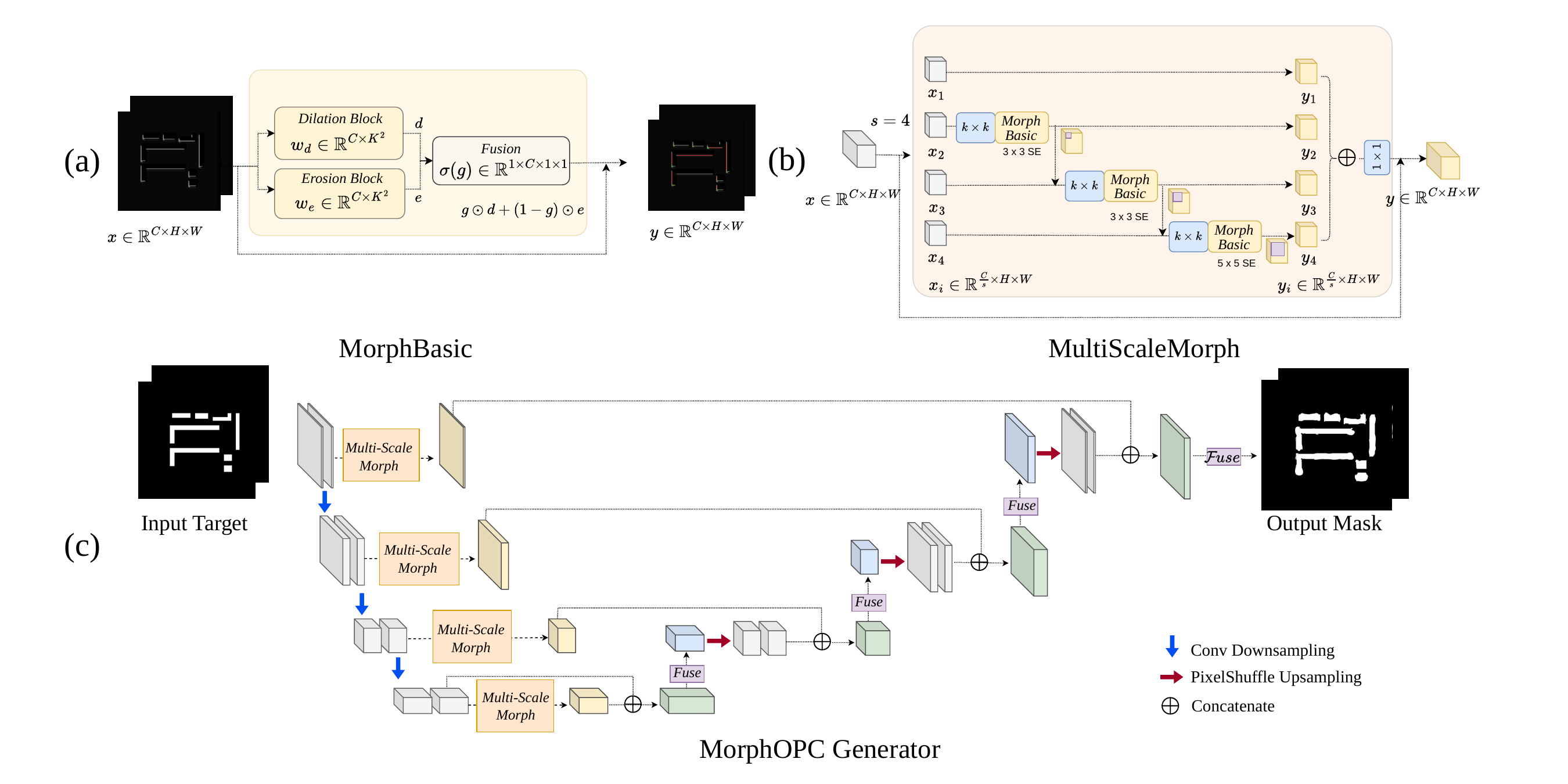}}
\caption{Illustration of MorphOPC architecture.}
\vspace{-3mm}
\label{fig2}
\end{figure*} 

\section{PRELIMINARIES}\label{sec2}
\subsection{Mask Optimization}
We denote the target layout as $Z_t$ and the printed image on the wafer as $Z$. Mask optimization adjusts the photomask $M$ to compensate for distortions between $Z_t$ and $Z$ arising from the optical proximity effect. Model-based OPC is widely used and operates by iteratively correcting specific layout features through forward lithography simulation.
The workflow involves segmenting each feature into small fragments, comparing the simulated printing contour $Z$ with the intended target $Z_t$, and applying localized corrections in which each fragment edge is displaced inward or outward according to the difference between $Z$ and $Z_t$. This iterative process continues until the corrected mask pattern produces a simulated wafer contour that closely matches the design specification \cite{ma2011pixel}. 

ML OPCs typically learn a direct mapping from the target layout \( Z_t \) to the optimized mask \( M \) with generative models:
\begin{equation}
f_{\theta}: Z_t \mapsto M.
\end{equation}
The training objective is minimizing the \(\ell_2\) loss between \( M \) and golden reference mask \( M^* \), as well as the discrepancy between the printed image \( Z \) simulated from \( M \) and the target \( Z_t \):
\begin{equation}
\mathcal{L}(\theta) = \| M - M^* \|_2^2 + \| Z - Z_t \|_2^2.
\end{equation}
For mask evaluation, there are standard metrics including: 

(1) \textit{\(\ell_2\) error}: \(\ell_2\) error evaluates the printing fidelity of the generated mask by measuring $\left\| Z - Z_t \right\|_2^2$. 

(2) \textit{EPE}: Edge placement error (EPE) refers to the vertical or horizontal misalignment, i.e., Manhattan distance from the lithography
contour of $Z$ to the desired contour of the target pattern $Z_t$.

(3) \textit{PVB}: Process variation band (PVB) is defined as the area between the outermost and innermost printed edges across all process conditions, reflecting the robustness of a mask to process variations, which is measured under ±2\% dose error and calculated as $\left| Z_{max} - Z_{min} \right|_2^2$, where $Z_{max}$ and $Z_{min}$ are the printed images under maximum and minimum process conditions. 

(4) \textit{Shot}: shot count denotes the number of rectangular shots for accurately replicating the mask shapes. 

\subsection{Mathematical Morphology}\label{classical_morph}
Mathematical Morphology is a set-theoretic framework for the analysis and manipulation of geometric structures in images \cite{shih2016automatic,shih2002threshold,shih2019development}. It provides a systematic methodology by treating an image as a set of points and prob the image geometry with a shape-dependent SE. Dilation $\oplus$ and erosion $\ominus$ are two operations in mathematical morphology. Let $f(x)$ represent the intensity value of the input image at coordinate $x$, $g(y)$ represent the value of the non-flat SE at offset $y$, and $D_g$ is a set of coordinates as the domain of $g$. 

\textbf{Dilation $\oplus$} is defined as the maximum of the sum of the reflected structuring element and the local neighborhood of the image:
\begin{equation}
    (f \oplus g)(x) = \sup_{y \in D_g} \{f(x-y) + g(y)\},
\end{equation}
Dilation expands regions and fills small holes or gaps in the image. It increases the size of objects by adding pixels to their boundaries, depending on the shape and size of the structuring element. 

\textbf{Erosion $\ominus$} is defined as the minimum of the difference between the image values and the structuring element values:
\begin{equation}
    (f \ominus g)(x) = \inf_{y \in D_g} \{f(x+y) - g(y)\}    
\end{equation}
Erosion shrinks bright regions and removes small bright artifacts or noise. It subtracts pixels from the object boundaries, effectively reducing the size of features.

As illustrated in Figure~\ref{fig1}, from a geometric perspective, OPC can be viewed as displacing the edge segments of target features inward or outward to compensate for lithographic distortions. Such edge-based modifications correspond directly to morphological operations, where erosion represents inward edge contraction, and dilation represents outward edge expansion. The final OPC mask can be interpreted as the compositional result of a sequence of different dilation and erosion operations applied to the local target features. Through appropriate combinations of dilation and erosion, morphological operators can capture and manipulate structural patterns that are highly relevant to mask optimization. This structural bias makes morphology particularly suitable for modeling the geometric corrections required in OPC, where the final mask shape depends not only on pixel intensities but also on the spatial configuration of layout features. However, manually designing structuring elements to capture diverse geometric configurations and selecting the appropriate morphological operations for each case is highly cumbersome and non-scalable \cite{zana2001segmentation,zhang2015improved}. To this end, we leverage neural morphological operators to automatically learn adaptive SEs and context-dependent combinations of dilation and erosion to model the OPC process.

\section{MorphOPC}
\subsection{Neural Morphological Operations}
To endow the generative OPC models with geometric reasoning, we first build the morphological block \textbf{MorphBasic} backboned with learnable non-flat morphological operators to process the output feature maps of convolutional modules. Unlike convolutional layers that aggregate local patterns through weighted summation, morphological operators perform \emph{max-plus} and \emph{min-plus} algebraic operations over local neighborhoods, enabling direct modeling of topological transformations such as edge shrinkage and expansion. Given an input feature map $\mathbf{x} \in \mathbb{R}^{B \times C \times H \times W}$, the learnable non-flat morphological dilation and erosion operations defined for each channel $c$ are as follows:
\begin{align}
(\mathbf{x} \oplus \mathbf{w})_c(p) &= \max_{q \in W}\,[\, \mathbf{x}_c(p - q) + \mathbf{w}_c(q)\,] + \beta_c, \label{eq:dilation} \\
(\mathbf{x} \ominus \mathbf{w})_c(p) &= \min_{q \in W}\,[\, \mathbf{x}_c(p + q) - \mathbf{w}_c(q)\,] + \beta_c, \label{eq:erosion}
\end{align}
where $p$ and $q$ are the pixel coordinates in the spatial domain. $p$ denotes the output pixel location being calculated, and $q$ denotes the relative position within the SE's window $W$. $\mathbf{w}_c(q)$ is a per-channel learnable continuous structuring surface that encodes spatial weights over the window $W$, and $\beta_c$ is a per-channel bias term. We assign learnable morphological operators to each feature channel, enabling the network to learn distinct geometric transformations tailored to various local design features. Intuitively, our neural morphological operators function similarly to pooling layers but with adaptive bias surfaces and are differentiable through the subgradients of the $\max$ and $\min$ functions.

To dynamically balance expansion and shrinkage, we fuse dilation and erosion outputs via a learnable per-channel gate:
\begin{equation}
\mathbf{y} = \sigma(\mathbf{g}) \odot (\mathbf{x} \oplus \mathbf{w}) 
          + \big[1 - \sigma(\mathbf{g})\big] \odot (\mathbf{x} \ominus \mathbf{w}),
\label{eq:morph-block}
\end{equation}
where $\sigma(\cdot)$ is the sigmoid activation and $\mathbf{g}$ is a channel-wise gating parameter. 
The gate controls whether each channel should emphasize dilation (feature expansion) or erosion (feature shrinkage), effectively learning a morphological operation prior. Each $MorphBasic$ block further includes batch normalization (BN), nonlinear activation $\phi(\cdot)$, and residual connection to the input:
\begin{equation}
\mathbf{y}_{MorphBasic} = \phi\!\left(\mathrm{BN}\!\big(\mathbf{P}(\mathbf{y} + \mathbf{x})\big)\right),
\end{equation}
where $\mathbf{P}$ is a $1 \times 1$ projection for linear combination.

\subsection{Multi-scale Hierarchical Morphological Learning}
Modern lithography-aware mask generation requires precise modeling of both local geometric primitives (e.g., edges, corners, and line-ends) and non-local pattern interactions induced by optical proximity effects. To effectively capture both fine local geometry and long-range pattern interactions, we propose a multi-scale hierarchical morphological module \textbf{MultiScaleMorph} that integrates conventional convolution with learnable morphological operators in a hierarchical residual framework, enabling structured multi-scale morphological learning.

Given the feature map $\mathbf{x} \in \mathbb{R}^{B \times C \times H \times W}$, we split it into $s$ groups along the channel dimension: $\mathbf{x} = [\mathbf{x}_1, \mathbf{x}_2, \dots, \mathbf{x}_s], \mathbf{x}_i \in \mathbb{R}^{B \times \frac{C}{s} \times H \times W}.$ Here, $s$ controls the scale dimension, larger $s$ allows features with richer
receptive field sizes to be processed by morphological operators. The first split $\mathbf{x}_1$ is preserved as a base representation, while subsequent splits are processed sequentially with MorphConv blocks:
\begin{equation}
\mathbf{y}_i =
\begin{cases}
\mathbf{x}_1, & i = 1, \\
MorphBasic_i\big( \mathcal{F}(\mathbf{x}_i) \big), & i = 2, \\
MorphBasic_i\big( \mathcal{F}(\mathbf{x}_i + \mathbf{y}_{i-1})), & 2 < i \le s,
\end{cases}
\end{equation}
where $\mathcal{F}(\cdot)$ denotes a standard convolutional block implemented as a $k \times k$ convolution, $MorphBasic_i\big( \mathcal{F}(\cdot))$ sequentially combines a convolutional operator with a basic learnable morphological block with scale-specific
configurations. Each feature split $\mathbf{x}_i$ is processed by a $k \times k$ convolution, which progressively enlarges its receptive field for subsequent neural morphological operations. Through the hierarchical residual connections, the module implicitly generates a combinatorial mixture of features with diverse receptive field sizes, enabling rich multi-scale representations.

To explicitly control the receptive field at different scales, each $MorphBasic_i$ employs distinct morphological kernel sizes, enabling progressive enlargement of the effective receptive field and capturing both fine-grained and long-range interactions. The multi-scale features are concatenated and projected back to the original channel dimension using a $1 \times 1$ convolution:
\begin{equation}
\mathbf{y}_{MultiScaleMorph} = \sigma \Big( \mathrm{BN} \big( \mathrm{Conv}_{1 \times 1}([\mathbf{y}_1, \dots, \mathbf{y}_s]) \big) + \mathbf{x} \Big).
\end{equation}

\subsection{MorphOPC Architecture}
As shown in Figure~\ref{fig2}, the proposed multi-scale hierarchical morphological modules are incorporated into an encoder-decoder structure for target-to-mask generation. Encoder progressively downsamples the input image, extracting hierarchical compressed feature maps that capture increasingly abstract representations. Decoder leverages these multi-scale features, which are augmented by skip connections from the corresponding encoder levels, to reconstruct high-resolution output that retains fine spatial detail \cite{cao2022swin,ruan2024vm,trebing2021smaat}. At each encoder scale $i$, convolutional layers extract features $\mathbf{f}_i$, which are then processed by \textbf{MultiScaleMorph} module to get the geometrically refined feature map $\mathbf{m}_i$:
\begin{equation}
\mathbf{f}_i = Conv(\mathbf{f}_{i-1}), \quad
\mathbf{m}_i = MultiScaleMorph(\mathbf{f}_i).
\end{equation}
In our design, each multi-scale morphological block adaptively refines local geometry by reshaping feature maps according to learned structuring surfaces. The decoder then reconstructs the mask features $\mathbf{u}_{j-1}$ by progressively upsampling and fusing with the corresponding morphological outputs:
\begin{equation}
\mathbf{u}_{j-1} = Up(Fuse([\mathbf{u}_{j},\,\mathbf{m}_{j}])),
\end{equation}
where $Up(\cdot)$ denotes a learnable upsampling operator (e.g., pixel-shuffle in our implementation), 
$[\cdot]$ represents feature concatenation, and $Fuse$ is a $3\times3$ convolutional fusion layer.  
This design ensures that reconstructed contours remain consistent with the geometric features captured in the encoder.
\begin{figure}[t]
\centerline{\includegraphics[width=8cm]{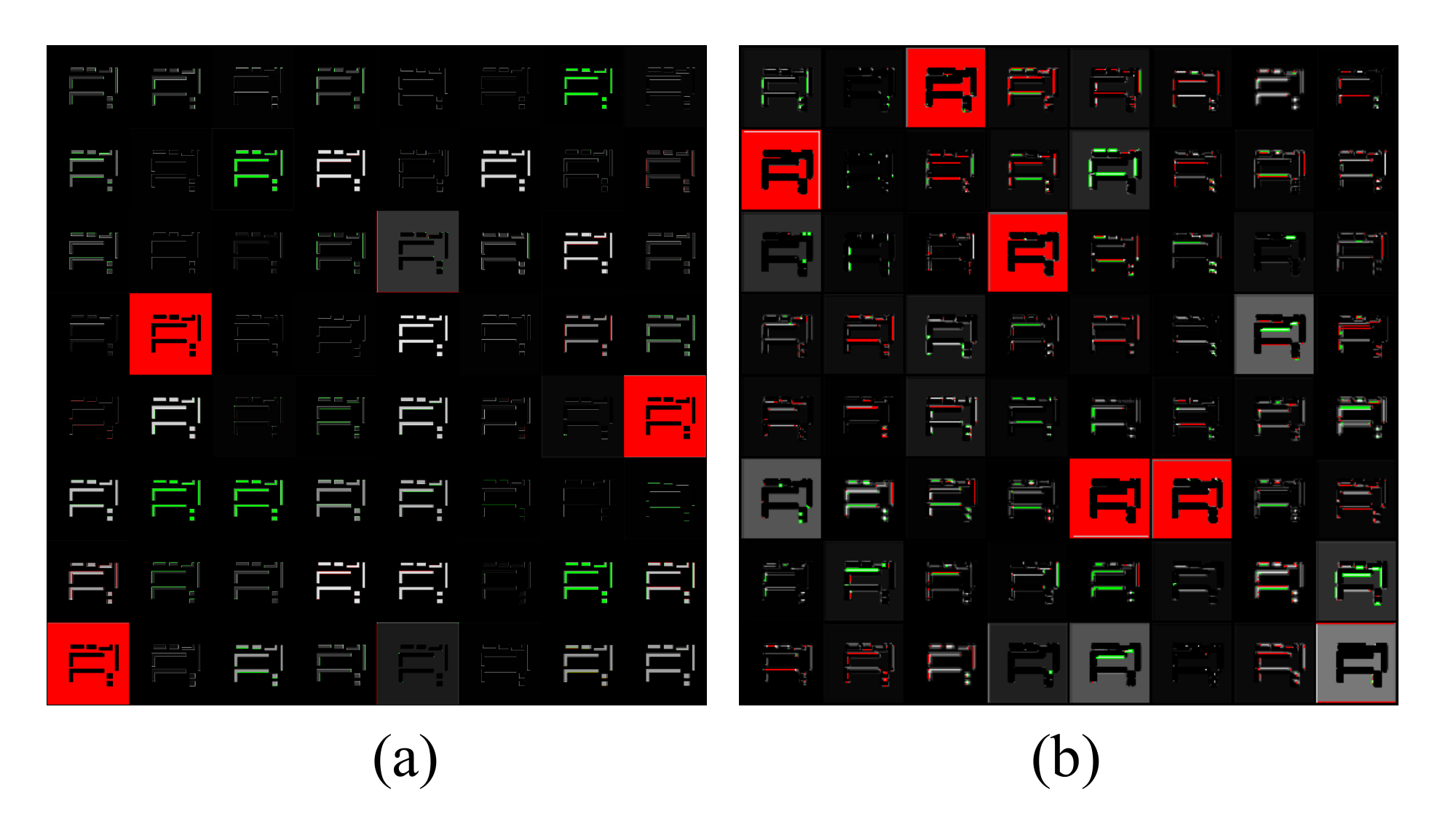}}
\caption{Views of morphological outputs relative to the input convolutional feature maps at (a) layer-2 and (b) layer-4.}
\label{morph}
\vspace{-3mm}
\end{figure}
\begin{table*}[t]
  \centering
  \caption{Comparison with learning-based models across OPC benchmarks.}
  \resizebox{\textwidth}{!}{
    \begin{tabular}{|c|cccc|cccc|cccc|cccc|cccc|}
      \hline
      \multirow{2}{*}{Model}
      & \multicolumn{4}{c|}{GAN-OPC}
      & \multicolumn{4}{c|}{LithoBench (MetalSet)}
      & \multicolumn{4}{c|}{LithoBench (ViaSet)}
      & \multicolumn{4}{c|}{MaskOpt (EBOPC)}
      & \multicolumn{4}{c|}{MaskOpt (ILT)} \\
      \cline{2-21}
      & \textit{\(\ell_2\)} & \textit{EPE} & \textit{PVB} & \textit{Shot}
      & \textit{\(\ell_2\)} & \textit{EPE} & \textit{PVB} & \textit{Shot}
      & \textit{\(\ell_2\)} & \textit{EPE} & \textit{PVB} & \textit{Shot}
      & \textit{\(\ell_2\)} & \textit{EPE} & \textit{PVB} & \textit{Shot}
      & \textit{\(\ell_2\)} & \textit{EPE} & \textit{PVB} & \textit{Shot} \\
      \hline
      GAN-OPC      & 8690 & 13.8 & 9183 & 633 & 39512 & 3.9 & 42230 & 490 & 21219 & \textbf{0.3} & \textbf{705} & 148 & 66012 & 16.6 & \textbf{6732} &  111 & 65158 & 0.4 & 8749 & 376 \\
      DAMO       & 8472 & 14.7 & \textbf{8949} & 862 & 33704 & 3.8 & 42306 &  532  & 5984 & 8.7 & 9899 & 180 & 58661 & 15.3 & 7072 & \textbf{96} & 58972 & 0.4 & 9082 & 404 \\
      NeuralILT  & 8658  & 14.2 & 10336 & 1392 & 38065 & 4.1 & 43753 & 489  & 7104 & 8.6 & 9546 & 281 & - & - & - & - & 48643 & 5.0 & \textbf{6749} & \textbf{187} \\
      CFNO       & 9020 & \textbf{13.1} & 10297 & \textbf{461} & 43443 & \textbf{3.6} & 44738 & \textbf{350} & 20325 & 0.7 & 2609  & \textbf{76} & - & - & - & - & 61968 & \textbf{0.3} & 8613 & 291 \\
      \cellcolor{yellow!10}\textbf{MorphOPC}
                & \cellcolor{yellow!10}\textbf{8107} & \cellcolor{yellow!10}13.6 & \cellcolor{yellow!10}9013 & \cellcolor{yellow!10}509 &  \cellcolor{yellow!10}\textbf{31915} & \cellcolor{yellow!10}4.1 & \cellcolor{yellow!10}\textbf{41300} & \cellcolor{yellow!10}499 &  \cellcolor{yellow!10}\textbf{4974} & \cellcolor{yellow!10}8.2 & \cellcolor{yellow!10}9549 & \cellcolor{yellow!10}274 &  \cellcolor{yellow!10}\textbf{53642} & \cellcolor{yellow!10}\textbf{4.9} & \cellcolor{yellow!10}8455 & \cellcolor{yellow!10}108
                & \cellcolor{yellow!10}\textbf{46860} & \cellcolor{yellow!10}4.0 & \cellcolor{yellow!10}6971 & \cellcolor{yellow!10}215  \\
      \hline
    \end{tabular}
  }
  \label{tab:tab1}
\end{table*}

\subsection{MorphOPC Training}
We adopt a two-stage training strategy consisting of a pretraining stage followed by a fine-tuning stage. In the pretraining stage, we train the MorphOPC generator to minimize the \textit{MSE} between the predicted mask and the ground-truth mask label. In the subsequent training stage, we adopt the GAN-style paradigm to minimize the MSE and l1loss of lithography printing. Specifically, we combine our generator $G$ with a discriminator $D$ under a GAN-based training paradigm to enhance both the realism and lithographic fidelity of the predicted mask patterns. During training, the generator produces a mask candidate $G(Z_t)$ from the target layout $Z_t$. This mask is then evaluated through a calibrated forward lithography simulator to obtain the printed image $Z$, enabling direct measurement of the lithographic discrepancy via the $\ell_2$ error between the printed pattern $Z$ and the target $Z_t$. The simulator output also allows comparison between the generated mask $G(Z_t)$ and the reference optimal mask $M^*$, capturing mask-level reconstruction accuracy.
To jointly enforce mask correctness and lithographic consistency, the generator is optimized using a composite loss that integrates mask reconstruction, printing error minimization, and an adversarial objective:
\begin{equation}
\begin{split}
\mathcal{L}_G = \min_{G} ( & \mathbb{E}_{Z_t \sim \mathcal{Z}}[\| M^* - G(Z_t) \|_2^2 + + \| Z - Z_t \|_2^2] \\
& - \mathbb{E}_{Z_t \sim \mathcal{Z}}[\log(D(Z_t, G(Z_t)))] ).v
\end{split}
\end{equation}
The discriminator $D$ is trained to maximize the probability of correctly classifying real masks ($M^*$) and generated masks ($G(Z_t)$). Its objective is to maximize:
\begin{equation}
\begin{split}
\mathcal{L}_D = \max_{D} \Big( & \mathbb{E}_{Z_t \sim \mathcal{Z}}[\log(D(M^*))] \\ 
& + \mathbb{E}_{Z_t \sim \mathcal{Z}}[\log(1 - D(G(Z_t)))] \Big).
\end{split}
\end{equation}
To showcase the geometric reasoning capability learned by MorphOPC, we visualize the outputs of the morphological modules at layers 2 and 4, as shown in Figure~\ref{morph}(a) and Figure~\ref{morph}(b). Each figure displays the 64 channel-wise outputs of the corresponding morphological block, where dilation-driven expansions are highlighted in green and erosion-driven contractions are highlighted in red. These visualizations reveal how the learned structuring elements modify the convolutional feature maps (e.g. the edge, line-end patterns of the input target) through localized geometric operations. As shown in the figure, the earlier morphological layer (layer 2) produces more coarse and spatially broad geometric adjustments, mainly capturing global expansions and contractions along dominant edges. In contrast, the deeper morphological layer (layer 4) exhibits significantly richer and more fine-grained transformations, selectively enhancing or suppressing intricate features. This progression indicates that deeper layers encode increasingly sophisticated local geometry, enabling MorphOPC to apply structurally adaptive corrections to layout patterns.

\section{Experiments}
\subsection{Benchmarks and Baselines}
We use three standard benchmark datasets to evaluate the performance of our method:  
(1) GAN-OPC \cite{yang2018gan}, which consists of 4{,}875 synthetic metal-layer tiles developed according to 32\,nm M1 layout design specifications; (2) LithoBench \cite{zheng2023lithobench}, which includes 16{,}472 synthesized tiles (MetalSet) for metal-layer ILT generated under ICCAD-13 design rules at the 32\,nm node and 116,415 clips (ViaSet) for via-layer ILT at the 45nm technology node with OpenROAD; and (3) MaskOpt \cite{hu2025maskopt}, which contains 104{,}714 metal-layer tiles for edge-based OPC (EBOPC) and ILT, built from real IC designs fabricated at the 45\,nm technology node. All images in GAN-OPC and LithoBench are $2048 \times 2048$ at 1\,pixel/nm$^2$, and MaskOpt images are $1024 \times 1024$ at 1\,pixel/nm$^2$ . For training and validation purposes, each dataset was split into an $80\%$ training set and a $20\%$ validation set. In the context of ML-based mask generation, we benchmark MorphOPC against representative state-of-the-art learning-based approaches, including GAN-OPC~\cite{yang2018gan}, DAMO~\cite{chen2020damo}, NeuralILT~\cite{jiang2020neural}, and CFNO~\cite{yang2022large}, across all datasets. The model sizes in number of trainable parameters are reported in parentheses: MorphOPC (44,581,377), GAN-OPC (20,559,652), DAMO (97,545,089), NeuralILT (7,787,905), and CFNO (1,867,073).

\subsection{Implementation Details}
We implement the proposed model using PyTorch 2.4.1. In the MultiScaleMorph module, the scale factor is set to $s = 8$ by default, where the corresponding morphological branches employ structuring elements determined by splitting the channels into two halves: the first half uses $3\times3$ structuring elements, while the second half uses $5\times5$ structuring elements. For the MorphBasic block, the per-channel gating parameters are initialized to zero, such that \(\sigma(g) \approx 0.5\), enabling a balanced fusion of dilation and erosion operations at initialization. Both the generator and discriminator were optimized using the Adam optimizer with a learning rate of \(1\times10^{-4}\), \(\beta_1 = 0.9\), and \(\beta_2 = 0.999\). The learning rate was decayed by a factor of \(0.1\) halfway through training, and models were trained for 10 epochs. We use 2 NVIDIA A100 GPUs with 80 GB VRAM each for the model training. The OpenILT\footnote{OPC Toolkit: \url{https://github.com/OpenOPC/OpenILT}} toolkit was used for both mask quality evaluation and forward lithography simulation. 
\begin{table*}[h!]
  \centering
  \caption{Test comparison on ICCAD 2013 contest benchmark.}
  \vspace{-3mm}
   \resizebox{\textwidth}{!}{
      \begin{tabular}{|c|cccc|cccc|cccc|cccc|cccc|}
      \hline
       \multicolumn{21}{|c|}{\cellcolor{gray!10}\textbf{ICCAD 2013 Benchmark (Pretrained on GAN-OPC)}} \\
       \hline
       \multirow{3}{*}{Test} & \multicolumn{4}{c|}{GAN-OPC} & \multicolumn{4}{c|}{DAMO} & \multicolumn{4}{c|}{NeuralILT} & \multicolumn{4}{c|}{CFNO} & \multicolumn{4}{c|}{\cellcolor{yellow!10}\textbf{MorphOPC}} \\
       \cline{2-21}
        & \textit{\(\ell_2\)} & \textit{EPE} & \textit{PVB} & \textit{shot} 
        & \textit{\(\ell_2\)} & \textit{EPE} & \textit{PVB} & \textit{Shot} 
        & \textit{\(\ell_2\)} & \textit{EPE} & \textit{PVB} & \textit{shot} 
        & \textit{\(\ell_2\)} & \textit{EPE} & \textit{PVB} & \textit{Shot} 
        & \cellcolor{yellow!10}\textit{\(\ell_2\)} & \cellcolor{yellow!10}\textit{EPE} & \cellcolor{yellow!10}\textit{PVB} & \cellcolor{yellow!10}\textit{Shot} \\
        \hline
        1  & 70120 & 9 & 55466 & 137 & 67407 & 8 & 49841 & 184 & 67628 & 10 & 64104 & 404 & 69332 & 6 & 49298 & 135 & \cellcolor{yellow!10}55356 & \cellcolor{yellow!10}10 & \cellcolor{yellow!10}51836 & \cellcolor{yellow!10}126  \\
        2  & 47234 & 8 & 46367 & 114 & 46291 & 8 & 46435 & 138 & 45027 & 7 & 52834 & 303 & 50620 & 5 & 53432 & 110 & \cellcolor{yellow!10}42015 & \cellcolor{yellow!10}8 & \cellcolor{yellow!10}46407 & \cellcolor{yellow!10}120  \\
        3  & 103656 & 6 & 85025 & 202 & 109048 & 6 & 67882 & 262 & 134856 & 9 & 74962 & 469 & 127178 & 0 & 74429 & 193 & \cellcolor{yellow!10}94007 & \cellcolor{yellow!10}1 & \cellcolor{yellow!10}64462 & \cellcolor{yellow!10}174  \\
        4  & 30165 & 0 & 24118 & 50 & 28385 & 1 & 24195 & 54 & 33372 & 2 & 36731 & 177 & 34783 & 1 & 36912 & 47 & \cellcolor{yellow!10}24585 & \cellcolor{yellow!10}1 & \cellcolor{yellow!10}23587 & \cellcolor{yellow!10}41  \\
        5  & 46856 & 0 & 58627 & 183 & 46451 & 0 & 53696 & 168 & 50904 & 0 & 58905 & 425 & 74344 & 0 & 68014 & 139 & \cellcolor{yellow!10}45686 & \cellcolor{yellow!10}0 & \cellcolor{yellow!10}57653 & \cellcolor{yellow!10}169  \\
        6  & 48158 & 2 & 53511 & 200 & 45826 & 2 & 48851 & 198 & 52477 & 2 & 57268 & 473 & 76936 & 1 & 68168 & 178 & \cellcolor{yellow!10}45378 & \cellcolor{yellow!10}2 & \cellcolor{yellow!10}52536 & \cellcolor{yellow!10}169   \\
        7  & 32535 & 0 & 48047 & 114 & 32909 & 0 & 44843 & 120 & 42645 & 0 & 48785 & 306 & 57411 & 0 & 65171 & 104 & \cellcolor{yellow!10}27177 & \cellcolor{yellow!10}0 & \cellcolor{yellow!10}45234 & \cellcolor{yellow!10}99  \\
        8  & 18792 & 0 & 23222 & 94 & 17592 & 0 & 22541 & 88 & 25596 & 0 & 25077 & 299 & 19817 & 0 & 25036 & 87 & \cellcolor{yellow!10}18346 & \cellcolor{yellow!10}0 & \cellcolor{yellow!10}23367 & \cellcolor{yellow!10}76  \\
        9  & 57084 & 1 & 62342 & 192 & 60842 & 2 & 58586 & 201 & 61612 & 1 & 67435 & 455 & 77852 & 1 & 81391 & 155 & \cellcolor{yellow!10}58890 & \cellcolor{yellow!10}1 & \cellcolor{yellow!10}60851 & \cellcolor{yellow!10}147  \\
        10 & 17435 & 0 & 19664 & 76 & 17762 & 0 & 19651 & 75 & 14472 & 0 & 18738 & 166 & 17724 & 0 & 19425 & 64 & \cellcolor{yellow!10}12329  & \cellcolor{yellow!10}0 & \cellcolor{yellow!10}19616 & \cellcolor{yellow!10}76  \\
        \hline   
        Avg.
            & 47204 & 2.6 & 47639 & 136
            & 47251 & 2.7 & 43652 & 149
            & 52859 & 3.1 & 50484 & 348 
            & 60600 & 1.4 & 54128 & 121 
            & \cellcolor{yellow!10}\textbf{42377} & \cellcolor{yellow!10}2.3 & \cellcolor{yellow!10}44555 & \cellcolor{yellow!10}\textbf{120} \\
        \hline  
      \end{tabular}
   }
   \resizebox{\textwidth}{!}{
      \begin{tabular}{|c|cccc|cccc|cccc|cccc|cccc|}
      \hline
       \multicolumn{21}{|c|}{\cellcolor{gray!10}\textbf{ICCAD 2013 Benchmark (Pretrained on LithoBench)}} \\
       \hline
       \multirow{3}{*}{Test} & \multicolumn{4}{c|}{GAN-OPC} & \multicolumn{4}{c|}{DAMO} & \multicolumn{4}{c|}{NeuralILT} & \multicolumn{4}{c|}{CFNO} & \multicolumn{4}{c|}{\cellcolor{yellow!10}\textbf{MorphOPC}} \\
       \cline{2-21}
        & \textit{\(\ell_2\)} & \textit{EPE} & \textit{PVB} & \textit{shot} 
        & \textit{\(\ell_2\)} & \textit{EPE} & \textit{PVB} & \textit{Shot} 
        & \textit{\(\ell_2\)} & \textit{EPE} & \textit{PVB} & \textit{shot} 
        & \textit{\(\ell_2\)} & \textit{EPE} & \textit{PVB} & \textit{Shot} 
        & \cellcolor{yellow!10}\textit{\(\ell_2\)} & \cellcolor{yellow!10}\textit{EPE} & \cellcolor{yellow!10}\textit{PVB} & \cellcolor{yellow!10}\textit{Shot} \\
        \hline
        1  & 49292 & 12 & 47803 & 535 & 48585 & 12 & 48537 & 533  & 47821 & 14 & 49721 & 496 & 55664 & 10 & 46107 & 373 & \cellcolor{yellow!10}44273 & \cellcolor{yellow!10}12 & \cellcolor{yellow!10}46998 & \cellcolor{yellow!10}492  \\
        2  & 38168 & 8 & 40593 & 478 & 35589 & 8 & 36093 & 522 & 40787 & 8 & 40696 & 405 & 40645 & 7 & 42522 & 299 & \cellcolor{yellow!10}34370 & \cellcolor{yellow!10}8 & \cellcolor{yellow!10}37162 & \cellcolor{yellow!10}448  \\
        3  & 83623 & 2 & 77500 & 602 & 81293 & 5 & 75069 & 671 & 81272 & 3 & 89678 & 579 & 97489 & 5 & 79691 & 444 & \cellcolor{yellow!10}74648 & \cellcolor{yellow!10}6 & \cellcolor{yellow!10}74259 & \cellcolor{yellow!10}554  \\
        4  & 15238 & 4 & 22763 & 470 & 10584 & 2 & 23214 & 432 & 18315 & 4 & 24197 & 343 & 19311 & 2 & 26064 & 201 & \cellcolor{yellow!10}11189 & \cellcolor{yellow!10}4 & \cellcolor{yellow!10}22560 & \cellcolor{yellow!10}444  \\
        5  & 46100 & 0 & 53467 & 506 & 36278 & 0 & 56138 & 614 & 42993 & 0 & 51357 & 585 & 50869 & 0 & 56445 & 441 & \cellcolor{yellow!10}37240 & \cellcolor{yellow!10}0 & \cellcolor{yellow!10}54021 & \cellcolor{yellow!10}562  \\
        6  & 45556 & 3 & 46289 & 581 & 35463 & 2 & 47603 & 646 & 42537 & 2 & 45958 & 590 & 49459 & 2 & 52558 & 477 & \cellcolor{yellow!10}35779 & \cellcolor{yellow!10}2 & \cellcolor{yellow!10}45826 & \cellcolor{yellow!10}638 \\
        7  & 29705 & 0 & 38633 & 440 & 23004 & 0 & 39659 & 528 & 23846 & 0 & 39232 & 527 & 31244 & 0 & 42200 & 321 & \cellcolor{yellow!10}17584 & \cellcolor{yellow!10}0 & \cellcolor{yellow!10}37933 & \cellcolor{yellow!10}543  \\
        8  & 18516 & 0 & 20640 & 421 & 15040 & 0 & 20807 & 439 & 16854 & 0 & 21722 & 446 & 20248 & 0 & 22269 & 293 & \cellcolor{yellow!10}13250 & \cellcolor{yellow!10}0 & \cellcolor{yellow!10}20298 & \cellcolor{yellow!10}454  \\
        9  & 52506 & 1 & 57649 & 509 & 41795 & 1 & 59190 & 636 & 51380 & 1 & 57266 & 633 & 55831 & 2 &  62400 & 446 & \cellcolor{yellow!10}40552 & \cellcolor{yellow!10}1 & \cellcolor{yellow!10}57316 & \cellcolor{yellow!10}557 \\
        10 & 16415 & 9 & 16967 & 355 & 9405 & 8 & 16746 & 299 & 14846 & 9 & 17707 & 283 & 13668 & 8 & 17124 & 201 & \cellcolor{yellow!10}10264 & \cellcolor{yellow!10}8 & \cellcolor{yellow!10}16623 & \cellcolor{yellow!10}296  \\
        \hline   
        Avg.
            & 39512 & 3.9 & 42230 & 490
            & 33704 & 3.8 & 42306 & 532 
            & 38065 & 4.1 & 43753 & 489 
            & 43443 & 3.6 & 44738 & 350 
            & \cellcolor{yellow!10}\textbf{31915} & \cellcolor{yellow!10}4.1 & \cellcolor{yellow!10}\textbf{41300} & \cellcolor{yellow!10}499  \\
        \hline  
      \end{tabular}
   }
   \vspace{-3mm}
  \label{tab:tab2}
\end{table*}

\begin{table*}[h!]
  \centering
  \caption{Test comparison on larger designs.}
  \vspace{-3mm}
  \resizebox{\textwidth}{!}{
    \begin{tabular}{|c|cccc|cccc|cccc|cccc|cccc|}
    \hline
    \multicolumn{21}{|c|}{\cellcolor{gray!10}\textbf{Larger Design Benchmark (Pretrained on GAN-OPC)}} \\
    \hline
    \multirow{2}{*}{Test} 
    & \multicolumn{4}{c|}{GAN-OPC} 
    & \multicolumn{4}{c|}{DAMO} 
    & \multicolumn{4}{c|}{NeuralILT} 
    & \multicolumn{4}{c|}{CFNO} 
    & \multicolumn{4}{c|}{\cellcolor{yellow!10}\textbf{MorphOPC}} \\
    \cline{2-21}
    & \textit{\(\ell_2\)} & \textit{EPE} & \textit{PVB} & \textit{shot} 
    & \textit{\(\ell_2\)} & \textit{EPE} & \textit{PVB} & \textit{shot} 
    & \textit{\(\ell_2\)} & \textit{EPE} & \textit{PVB} & \textit{shot} 
    & \textit{\(\ell_2\)} & \textit{EPE} & \textit{PVB} & \textit{shot}  
    & \cellcolor{yellow!10}\textit{\(\ell_2\)} & \cellcolor{yellow!10}\textit{EPE} & \cellcolor{yellow!10}\textit{PVB} & \cellcolor{yellow!10}\textit{shot}  \\
    \hline
    11 & 110163 & 18 & 106714 & 311 & 96853 & 15 & 104298 & 340 & 108584 & 20 & 112203 & 749 & 124190 & 13 & 111745 & 259 & \cellcolor{yellow!10}94579 & \cellcolor{yellow!10}13 & \cellcolor{yellow!10}109699 & \cellcolor{yellow!10}283 \\
    12 & 104544 & 13 & 104633 & 288 & 83858 & 11 & 99311 & 282 & 89455 & 12 & 109796 & 551 & 103606 & 9 & 123921 & 258 & \cellcolor{yellow!10}79698 & \cellcolor{yellow!10}12 & \cellcolor{yellow!10}107312 & \cellcolor{yellow!10}237 \\
    13 & 159272 & 17 & 132406 & 346 & 147885 & 15 & 125778 & 404 & 142771 & 17 & 135333 & 821 & 177046 & 11 & 141396 & 304 & \cellcolor{yellow!10}134747 & \cellcolor{yellow!10}12 & \cellcolor{yellow!10}138711 & \cellcolor{yellow!10}314 \\
    14 & 86750 & 4 & 81864 & 196 & 65697 & 3 & 80406 & 215 & 68567 & 5 & 92746 & 499 & 83450 & 3 & 104105 & 168 & \cellcolor{yellow!10}62043 & \cellcolor{yellow!10}5 & \cellcolor{yellow!10}81967 & \cellcolor{yellow!10}196 \\
    15 & 98851 & 14 & 115064 & 340 & 92335 & 12 & 105305 & 289 & 101138 & 13 & 120607 & 665 & 131963 & 10 & 122759 & 300 & \cellcolor{yellow!10}90114 & \cellcolor{yellow!10}11 & \cellcolor{yellow!10}107401 & \cellcolor{yellow!10}294 \\
    16 & 96882 & 15 & 110031 & 295 & 91881 & 13 & 101776 & 335 & 96208 & 9 & 117692 & 688 & 142813 & 7 & 129182 & 296 & \cellcolor{yellow!10}87032 & \cellcolor{yellow!10}13 & \cellcolor{yellow!10}111938 & \cellcolor{yellow!10}279 \\
    17 & 72726 & 3 & 87928 & 244 & 57230 & 2 & 84798 & 256  & 61456 & 3 & 93525 & 593 & 79895 & 4 & 101490 & 228 & \cellcolor{yellow!10}61610 & \cellcolor{yellow!10}3 & \cellcolor{yellow!10}89957 & \cellcolor{yellow!10}236 \\
    18 & 72015 & 7 & 81157 & 260 & 56252 & 6 & 79071 & 248  & 62441 & 7 & 82075 & 615 & 80735 & 6 & 94798 & 227 & \cellcolor{yellow!10}56979 & \cellcolor{yellow!10}7 & \cellcolor{yellow!10}81951 & \cellcolor{yellow!10}248 \\
    19 & 106914 & 24 & 114137 & 394 & 126929 & 23 & 108583 & 341 & 104953 & 23 & 132139 & 687 & 150281 & 11 & 136064 & 322 & \cellcolor{yellow!10}100414 & \cellcolor{yellow!10}17 & \cellcolor{yellow!10}112244 & \cellcolor{yellow!10}330 \\
    20 & 69677 & 12 & 77272 & 234 & 52183 & 10 & 75311 & 226 & 54880 & 12 & 75541 & 504 & 72029 & 11 & 90908 & 195 & \cellcolor{yellow!10}48890 & \cellcolor{yellow!10}12 & \cellcolor{yellow!10}77391 & \cellcolor{yellow!10}231 \\
    \hline
    Avg.
    & 97779 & 12.7 & 101121 & 291
    & 87110 & 11.0 & 96464 & 294
    & 89045 & 12.1 & 107166 & 637 
    & 114601 & 8.5 & 115637 & 256
    & \cellcolor{yellow!10}\textbf{81611} & \cellcolor{yellow!10}10.5  & \cellcolor{yellow!10}101857 & \cellcolor{yellow!10}265 \\
    \hline
    \end{tabular}
    }
    \resizebox{\textwidth}{!}{
    \begin{tabular}{|c|cccc|cccc|cccc|cccc|cccc|}
    \hline
    \multicolumn{21}{|c|}{\cellcolor{gray!10}\textbf{Larger Design Benchmark (Pretrained on Lithobench)}} \\
    \hline
    \multirow{2}{*}{Test} 
    & \multicolumn{4}{c|}{GAN-OPC} 
    & \multicolumn{4}{c|}{DAMO} 
    & \multicolumn{4}{c|}{NeuralILT} 
    & \multicolumn{4}{c|}{CFNO} 
    & \multicolumn{4}{c|}{\cellcolor{yellow!10}\textbf{MorphOPC}} \\
    \cline{2-21}
    & \textit{\(\ell_2\)} & \textit{EPE} & \textit{PVB} & \textit{shot} 
    & \textit{\(\ell_2\)} & \textit{EPE} & \textit{PVB} & \textit{shot} 
    & \textit{\(\ell_2\)} & \textit{EPE} & \textit{PVB} & \textit{shot} 
    & \textit{\(\ell_2\)} & \textit{EPE} & \textit{PVB} & \textit{shot}  
    & \cellcolor{yellow!10}\textit{\(\ell_2\)} & \cellcolor{yellow!10}\textit{EPE} & \cellcolor{yellow!10}\textit{PVB} & \cellcolor{yellow!10}\textit{shot}  \\
    \hline
    11 & 89266 & 19 & 97858 & 749 & 81234 & 21 & 98234 & 772 & 88042 & 22 & 95043 & 858 & 94613 & 18 & 93406 & 700 & \cellcolor{yellow!10}75173 & \cellcolor{yellow!10}20 & \cellcolor{yellow!10}95449 & \cellcolor{yellow!10}757   \\
    12 & 74611 & 17 & 90759 & 722 & 74635 & 16 & 94365 & 761 & 77265 & 17 & 90148 & 766 & 83201 & 10 & 91136 & 594 & \cellcolor{yellow!10}63360 & \cellcolor{yellow!10}19 & \cellcolor{yellow!10}89053  &  \cellcolor{yellow!10}724  \\
    13 & 126670 & 11 & 127975 & 830 & 118615 & 12 & 130669 & 866 & 124900 & 12 & 130996 & 931&135418 & 18 & 129456 & 727 & \cellcolor{yellow!10}112007 & \cellcolor{yellow!10}13  & \cellcolor{yellow!10}124328 & \cellcolor{yellow!10}789 \\
    14 & 51217 & 7 & 70138 & 734 & 46389 & 6 & 75514 & 762 & 56631 & 8 & 70896 & 706 &60562 & 4 & 75348 & 536 & \cellcolor{yellow!10}45080 & \cellcolor{yellow!10}5 & \cellcolor{yellow!10}74123 & \cellcolor{yellow!10}737 \\
    15 & 87375 & 20 & 104314 & 698 & 65063 & 19 & 106124 & 791 & 86440 & 21 & 100224 & 849 & 114561 & 16 & 120293 & 648 & \cellcolor{yellow!10}69119 & \cellcolor{yellow!10}21 & \cellcolor{yellow!10}102393 & \cellcolor{yellow!10}781 \\
    16 & 87155 & 13 & 99354 & 753 & 69556 & 11 & 103974 & 749 & 83633 & 15 & 95298 & 866 & 102130 & 9 & 107011 & 628 & \cellcolor{yellow!10}66745 & \cellcolor{yellow!10}14 & \cellcolor{yellow!10}98227 & \cellcolor{yellow!10}715 \\
    17 & 64298 & 10 & 86176 & 640 & 46878 & 9 & 81323 & 679 & 48876 & 9 & 78477 & 725 & 63350 & 9 & 80769 & 563 & \cellcolor{yellow!10}39887 & \cellcolor{yellow!10}9 & \cellcolor{yellow!10}77105  & \cellcolor{yellow!10}666  \\
    18 & 54240 & 11 & 68204 & 696 & 48299 & 11 & 72014 & 723 & 56715 & 12 & 68146 & 816 & 62623 & 11 & 70236 & 649 & \cellcolor{yellow!10}43424 & \cellcolor{yellow!10}12 &  \cellcolor{yellow!10}68773 & \cellcolor{yellow!10}744  \\
    19 & 95727 & 26 & 111820 & 722 & 75131 & 24 & 113190 & 790 & 99088 & 22 & 105357 & 862 &133442 & 20 & 115046 & 642 & \cellcolor{yellow!10}78526 & \cellcolor{yellow!10}25 & \cellcolor{yellow!10}110380 & \cellcolor{yellow!10}769  \\
    20 & 45425 & 12 & 64390 & 663 & 43597 & 12 & 68564 & 691 & 54920 & 13 & 63984 & 712 & 61218 & 12 & 67205 & 594 & \cellcolor{yellow!10}39364 & \cellcolor{yellow!10}12 & \cellcolor{yellow!10}64815 & \cellcolor{yellow!10}677 \\
    \hline
    Avg.
    & 77598 & 14.6 & 92099  &  721
    & 66940 & 14.1 & 94397 & 758
    & 77651 & 15.1  & 89857  & 809
    & 91112 & 12.7 & 94991 & 628
    & \cellcolor{yellow!10}\textbf{63268} & \cellcolor{yellow!10}15.0 & \cellcolor{yellow!10}90465 & \cellcolor{yellow!10}736 \\
    \hline
    \end{tabular}
    }
    \vspace{-3mm}
    \label{tab:larger}
\end{table*}
\begin{figure*}[h!]
\centerline{\includegraphics[width=15cm]{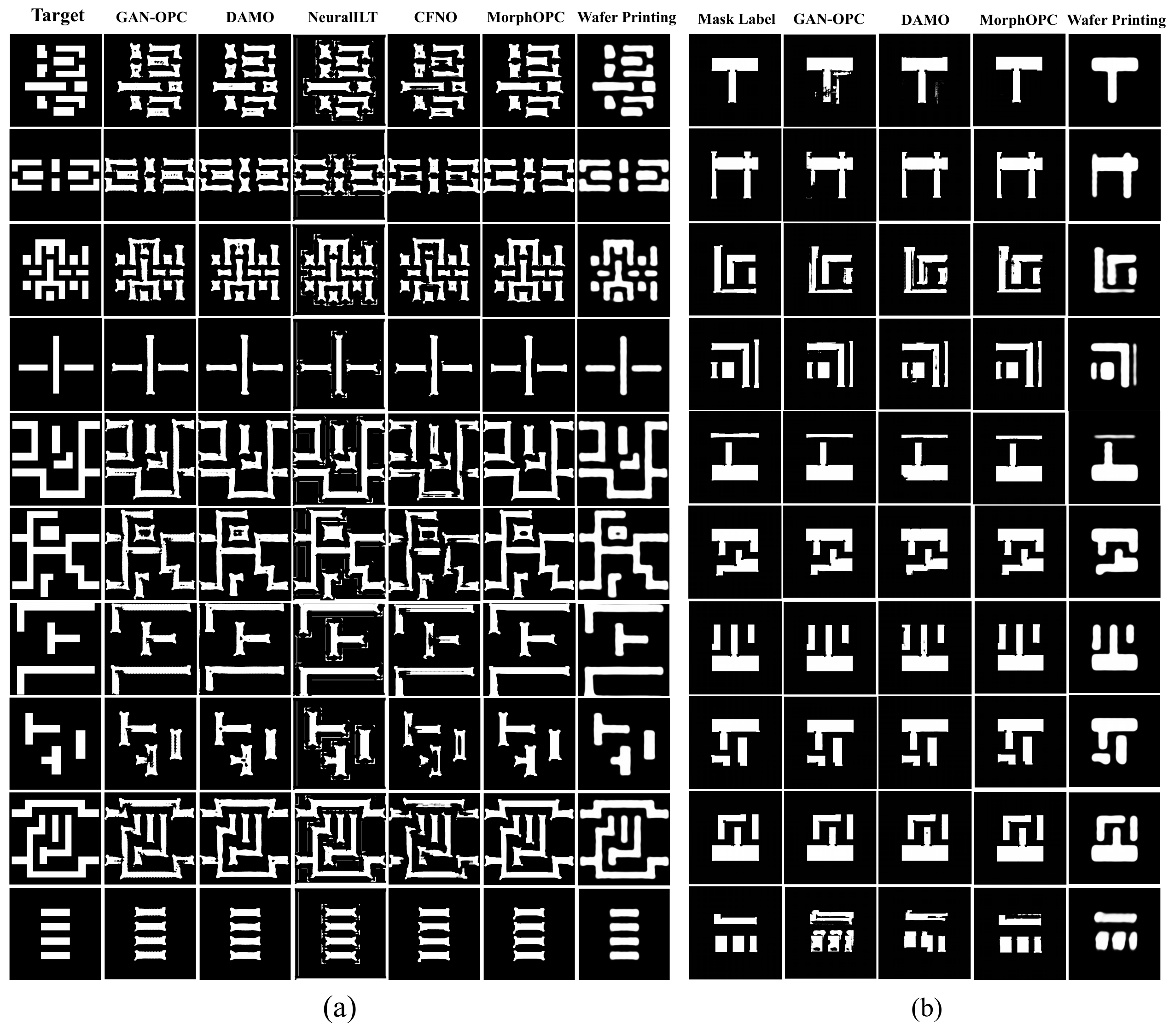}}
\vspace{-3mm}
\caption{Test result visualization. Rows correspond to ten test cases from (a) the ICCAD 2013 benchmark, and (b) the MaskOpt edge-based OPC set: from top to bottom, each row corresponds to tiles of different standard cells, including AOI21\_X2, BUF\_X16, OAI221\_X2, OR4\_X2, BUF\_X32, NAND3\_X4, NOR3\_X4, AND3\_X1, CLKBUF\_X1, and OR3\_X1.}
\label{fig3}
\vspace{-3mm}
\end{figure*}

\subsection{Experimental Results}
\noindent\textbf{Main Results.}
We first compare the proposed model with ML baselines across all benchmarks in Table~\ref{tab:tab1}. We report \textit{\(\ell_2\)} error, \textit{EPE}, \textit{PVB}, and \textit{Shot}, as defined in Section~\ref{sec2}. Lower values correspond to better performance across all metrics. All models are trained on their respective datasets. On the GAN-OPC dataset, MorphOPC attains the lowest \textit{\(\ell_2\)} error. Although CFNO achieves the best \textit{EPE} and \textit{Shot}, its \textit{\(\ell_2\)} error is 11.3\% higher than that of MorphOPC, which is prohibitively large and undermines mask usability. Similarly, while DAMO achieves the lowest \textit{PVB}, its \textit{\(\ell_2\)} error is 4.5\% higher, and its \textit{EPE} is 8.1\% higher than MorphOPC, indicating an overall inferior trade-off in mask quality. On the MetalSet of LithoBench, MorphOPC achieves the lowest \textit{\(\ell_2\)} error and \textit{PVB}. While CFNO obtains the best \textit{EPE} and \textit{Shot}, it suffers from the worst \textit{\(\ell_2\)} error and \textit{PVB}. On the ViaSet of LithoBench, MorphOPC achieves the lowest \textit{\(\ell_2\)} error. Although GAN-OPC achieves the best \textit{EPE} and \textit{PVB}, its \textit{\(\ell_2\)} error is more than 4$\times$ higher than that of MorphOPC, indicating severely degraded mask fidelity. Furthermore, CFNO attains the lowest \textit{Shot}, but its \textit{\(\ell_2\)} error remains over 4$\times$ higher, reinforcing the superiority of MorphOPC in balancing accuracy and manufacturability. For edge-based OPC mask generation on MaskOpt, we include GAN-OPC and DAMO as baselines, as they are capable of generating characteristic features such as serifs. NeuralILT and CFNO are excluded since they are inherently designed for ILT masks. From the results of EBOPC mask generation, MorphOPC achieves the best overall performance in terms of \textit{\(\ell_2\)} error and \textit{EPE}. Specifically, it reduces the \textit{\(\ell_2\)} error by 18.7\% compared to GAN-OPC and 8.5\% compared to DAMO, while significantly lowering EPE by 70.5\% and 68.0\%, respectively. For the ILT mask generation on MaskOpt dataset, MorphOPC continues to demonstrate strong performance by achieving the lowest \textit{\(\ell_2\)} error, corresponding to a 3.6\% reduction compared to the competing baseline NeuralILT. It also achieves a lower \textit{EPE} than NeuralILT, further highlighting its advantage in printing fidelity. It maintains competitive \textit{PVB} and significantly reduces \textit{Shot} count by over 40\% compared to GAN-OPC and DAMO, indicating improved manufacturability and lower mask complexity. Overall, MorphOPC achieves a favorable balance between mask fidelity and manufacturability, consistently delivering low \textit{\(\ell_2\)} error while maintaining competitive \textit{EPE}, \textit{PVB}, and reduced \textit{Shot} count. This demonstrates its practical applicability as an efficient ML-based surrogate for traditional OPC methods in large-scale design flows. Although MorphOPC is primarily designed for edge-based OPC, the results indicate that it adapts effectively to both EBOPC and ILT tasks across metal and via layers. \\

\noindent\textbf{Test on ICCAD 2013 Benchmark.} To evaluate the generalization capability of MorphOPC and the baseline models on unseen layouts, we conduct experiments on the ICCAD 2013 benchmark~\cite{banerjee2013iccad}, as summarized in Table~\ref{tab:tab2}. Specifically, models pretrained on the GAN-OPC and LithoBench datasets are directly evaluated on ICCAD 2013, as both datasets are derived from 32\,nm technology node, enabling a fair assessment of cross-design generalization. For models pretrained on GAN-OPC, MorphOPC achieves the highest printing fidelity, attaining the lowest \textit{\(\ell_2\)} error and \textit{Shot}, while maintaining a \textit{PVB} comparable to the best-performing baseline DAMO. These results demonstrate strong generalization capability across unseen layouts. In contrast, although CFNO achieves the lowest \textit{EPE}, its \textit{\(\ell_2\)} error is 43.0\% higher than that of MorphOPC, indicating inferior overall mask accuracy. For models pretrained on LithoBench, MorphOPC consistently outperforms all baselines by achieving the lowest \textit{\(\ell_2\)} error and \textit{PVB}, demonstrating robust generalization under more complex training distributions. While CFNO attains the lowest \textit{Shot}, it exhibits a 36.1\% higher \textit{\(\ell_2\)} error compared to MorphOPC. Similarly, although DAMO achieves slightly better \textit{EPE}, its \textit{\(\ell_2\)} error remains 5.6\% higher and its \textit{PVB} is 2.4\% higher, indicating a suboptimal trade-off in mask fidelity. Overall, these results further validate that MorphOPC generalizes more effectively across datasets and maintains superior accuracy–efficiency balance. \\

\noindent\textbf{Test on Larger Designs.} To further validate the robustness of our model, we conduct experiments on a larger dataset and compare its performance with baselines in Table~\ref{tab:larger}. Under both pretraining settings, MorphOPC consistently achieves superior performance. When pretrained on GAN-OPC, MorphOPC achieves the lowest \textit{$\ell_2$} error and \textit{shot} count, while maintaining competitive \textit{EPE} and \textit{PVB}. Compared to the strong baseline DAMO, MorphOPC reduces the \textit{$\ell_2$} error by 6.3\% and further improves \textit{EPE}. When pretrained on LithoBench, MorphOPC also outperforms all baselines in \textit{$\ell_2$} error. While its \textit{EPE} is comparable to other learning-based approaches, MorphOPC maintains a favorable balance between mask quality and manufacturability, as reflected by competitive \textit{PVB} and moderate shot count. \\

\noindent\textbf{Results Visualization.} We visualize the masks generated by MorphOPC and SOTA baselines in Figure~\ref{fig3}. On the ICCAD 2013 benchmark, MorphOPC consistently produces cleaner and more complete mask patterns across all test cases compared to competing methods. While other models perform well on simpler patterns, their performance degrades significantly on more complex test cases. For the edge-based OPC examples from MaskOpt, MorphOPC demonstrates more fine-grained corrections on real IC design patterns, particularly exhibiting clearer serif structures at L-shaped corners. These results highlight the strong robustness of \textit{MorphOPC}, enabling effective adaptation to diverse geometries and maintaining high predictive consistency on industrial benchmark layouts.
\begin{figure}[t]
\centerline{\includegraphics[width=7cm]{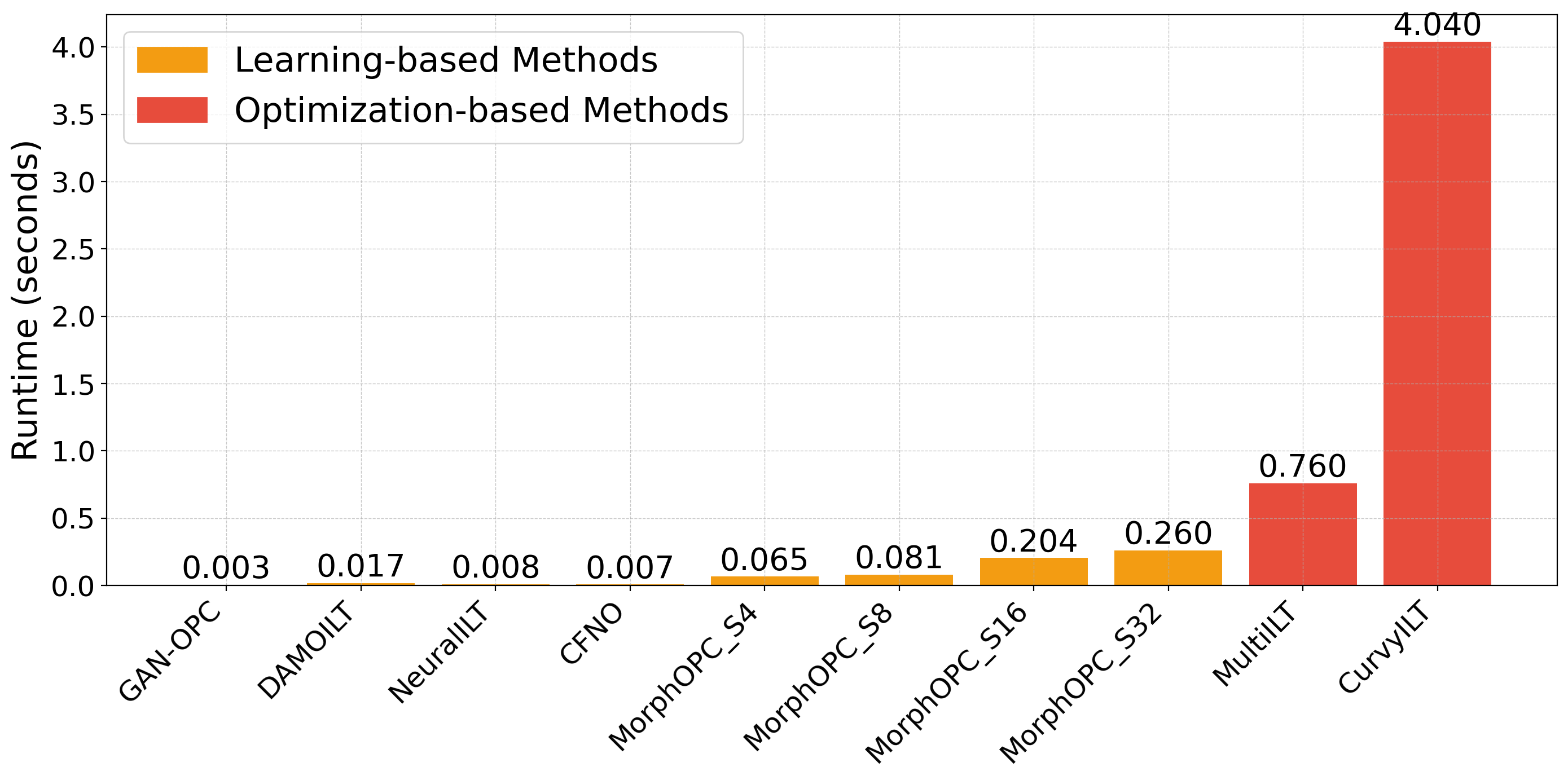}}
\caption{Average running time on ICCAD 2013 benchmark.}
\label{runtime}
\vspace{-5mm}
\end{figure}
\subsection{Analysis and Discussion} 

\noindent\textbf{Ablation Study.} We perform ablation studies to quantify the contributions of the \textit{MorphBasic} module and the hierarchical \textit{MultiScaleMorph} design. Case 1 preserves the hierarchical structure but replaces \textit{MorphBasic} modules with standard convolutions at each scale, effectively reducing the encoder to a hierarchical convolutional neural network. Case 2 removes the multi-scale hierarchy by applying only a \textit{MorphBasic} module after initial convolution, without channel splitting and hierarchical receptive field expansion. Case 3 uses a conventional U-Net as the mask generator. All ablation models are trained on the GAN-OPC dataset. We report their performance on both the GAN-OPC validation set and the ICCAD 2013 benchmark in Table~\ref{tab:ablation}. The results show that removing either the \textit{MorphBasic} module or the hierarchical \textit{MultiScaleMorph} design leads to consistent performance degradation. Case 1 increases the \textit{$\ell_2$} error by 0.69\% on GAN-OPC and raises the ICCAD \textit{$\ell_2$} error by 13.2\%, indicating that standard convolutions fail to capture the geometric transformations required for accurate OPC mask generation. Case 2 results in similar degradation, with the \textit{$\ell_2$} error increasing by 3.66\% on GAN-OPC and by 12.4\% on ICCAD, demonstrating the importance of multi-scale receptive fields for handling layout patterns with varying spatial complexity. When both components are removed in Case 3, the performance degrades most significantly: the GAN-OPC \textit{$\ell_2$} error increases by 7.1\% and the ICCAD \textit{$\ell_2$} error increases by 11.6\%, accompanied by a notable increase in \textit{EPE}.

\vspace{2mm}
\noindent\textbf{Scaling Factor Choice.} We analyze the choice of scale facor $s$ on the model performance in Table~\ref{tab:scale}. In MorphOPC architecture, a higher scale factor $s$ means that the feature channels are divided into more smaller subsets, enabling a finer-grained multi-scale representation within a single residual block to improve the representation capacity. From the results, we observe that a moderate scale factor achieves the best overall performance. Specifically, $s=8$ yields the lowest \textit{EPE} and \textit{PVB} on the GAN-OPC dataset, as well as the lowest \textit{$\ell_2$} error on the ICCAD 2013 benchmark. Increasing the scale factor further to $s=16$ reduces \textit{$\ell_2$} error on the GAN-OPC and \textit{EPE} on ICCAD, but leads to degradation in other metrics, suggesting diminishing returns from overly fine channel partitioning. When $s$ is increased to 32, the performance degrades significantly on most metrics. Overall, these results demonstrate that while multi-scale decomposition is beneficial, excessively large scale factors may harm feature integration across channels, and $s=8$ provides the most effective trade-off for the proposed architecture.

\begin{table}[t]
  \centering
  \caption{Ablation Results.}
    \vspace{-3mm}
  \resizebox{\columnwidth}{!}
  {
  \begin{tabular}{|c|ccccc|cccc|}
    \hline
    \multirow{2}{*}{Ablation Experiment} & \multicolumn{5}{c|}{GAN-OPC} & 
    \multicolumn{4}{c|}{ICCAD 2013 Benchmark} \\  
    \cline{2-10}
     & \textit{MSE} & \textit{\(\ell_2\)} & \textit{EPE} & \textit{PVB} & \textit{Shot} & \textit{\(\ell_2\)} & \textit{EPE} & \textit{PVB} & \textit{Shot} \\
    \hline
    \cellcolor{yellow!10}MorphOPC  & \cellcolor{yellow!10}\textbf{0.0046} & \cellcolor{yellow!10}\textbf{8107} & \cellcolor{yellow!10}\textbf{13.6} & \cellcolor{yellow!10}\textbf{9013} & \cellcolor{yellow!10}\textbf{509} & \cellcolor{yellow!10}\textbf{42377} & \cellcolor{yellow!10}\textbf{2.3} & \cellcolor{yellow!10}44555 & \cellcolor{yellow!10}\textbf{119.7}   \\
    \hline
    \makecell{C1. \textit{MorphBasic} \xmark}   & 0.0047 & 8163 & 13.9 & 9176 & 514 & 47969 & 2.4 & 45484 & 126.1 \\
    \hline
    \makecell{C2. \textit{MultiScaleMorph} \xmark}   & 0.0052 & 8404 & 13.9 & 9233 & 528 & 47651 & 2.3 & 50271 & 121.0 \\
    \hline
    \makecell{C3. \textit{MorphBasic,MultiScaleMorph} \xmark}   & 0.0056 & 8679 & 14.7 & 10571 & 678 & 47304 & 2.7 & \textbf{43679} & 151.2  \\
    \hline
  \end{tabular}
  }
  \vspace{-3mm}
  \label{tab:ablation}
\end{table}
\begin{table}[t]
  \centering
  \caption{Impact of scale factor on model performance.}
    \vspace{-3mm}
  \resizebox{\columnwidth}{!}
  {
  \begin{tabular}{|c|ccccc|cccc|}
    \hline
    \multirow{2}{*}{\makecell{MorphOPC \\ Scale Factor}} & \multicolumn{5}{c|}{GAN-OPC} & \multicolumn{4}{c|}{ICCAD 2013} \\
    \cline{2-10}
     & \textit{MSE} & \textit{\(\ell_2\)} & \textit{EPE} & \textit{PVB} & \textit{Shot} & \textit{\(\ell_2\)} & \textit{EPE} & \textit{PVB} & \textit{Shot} \\
    \hline
    $s=4$   & 0.0049 & 8410 & 14.4  & 9312 & \textbf{506} & 44251 & 2.3 & 44554 & 123.3 \\
    $s=8$   & \textbf{0.0046} & 8107 & \textbf{13.6} & \textbf{9013} & 509 & \textbf{42377} & 2.3  & 44555 & 119.7\\
    $s=16$   & 0.0052 & \textbf{7877} & 14.4 & 9385 & 530 & 44161 & \textbf{1.7} & 47761 & 113.7 \\
    $s=32$   & 0.0068 & 8833 & 14.0 & 9234 & 753 & 53634 & 1.9 & \textbf{44442} & \textbf{97.8}  \\
    \hline
  \end{tabular}
  }
  \label{tab:scale}
  \vspace{-3mm}
\end{table}
\begin{table}[t]
  \centering
  \caption{Initialize optimization-based OPC with model-generated mask on ICCAD 2013 benchmark.}
    \vspace{-3mm}
  \resizebox{\columnwidth}{!}
  {
  \begin{tabular}{|l|ccccc|ccccc|}
    \hline
    \multirow{2}{*}{Model} & \multicolumn{5}{c|}{MultiILT} & \multicolumn{5}{c|}{CurvyILT} \\
    \cline{2-11}
     & \textit{\(\ell_2\)} & \textit{EPE} & \textit{PVB} & \textit{Shot} & \textit{Slowdown} & \textit{\(\ell_2\)} & \textit{EPE} & \textit{PVB} & \textit{Shot} & \textit{Speedup} \\
    \hline
    GAN-OPC & 29239 & 3.6 & 44501 & 527 & 2.04x & 27865 & 3.6 & 42657 & 526 & 3.97\% \\
    DAMO & 29165 & 3.6 & 44423 & 531 & 2.00x & 27753 & 3.6 & 42654 & 526 & 4.29\% \\
    NeuralILT & 28927 & 3.6 & 44666 & 507 & 2.00x & 27685 & 3.6 & 42919 & 516 & 4.44\% \\
    CFNO & 29266 & 3.6 & 44696 & 490 & 1.99× & 27813 & 3.6 & 42582 & 492 &  4.21\% \\
    \cellcolor{yellow!10}MorphOPC & \cellcolor{yellow!10}\textbf{28902} & \cellcolor{yellow!10}3.6 & \cellcolor{yellow!10}\textbf{44309} & \cellcolor{yellow!10}527 & \cellcolor{yellow!10}\textbf{1.98×} & \cellcolor{yellow!10}\textbf{27501} & \cellcolor{yellow!10}3.6 & \cellcolor{yellow!10}\textbf{42545} & \cellcolor{yellow!10}522 & \cellcolor{yellow!10}\textbf{4.52\%} \\
    \hline
  \end{tabular}
  }
  \label{tab:finetune}
  \vspace{-3mm}
\end{table}
\vspace{2mm}
\noindent\textbf{Running Time Analysis.} In Figure~\ref{runtime}, we compare the running time of MorphOPC and baselines against advanced optimization-based OPCs. We include MultiILT and CurvyILT as representative optimization-based OPC approaches due to their public availability. The results show that all learning-based methods achieve significantly faster inference compared to optimization-based OPCs. In particular, MorphOPC maintains a runtime on the order of $10^{-2}$ to $10^{-1}$ seconds, which is comparable to other ML baselines while delivering superior mask quality. Among different configurations, increasing the scale factor $s$ leads to a moderate increase in runtime. In contrast, optimization-based methods exhibit substantially higher runtimes. MultiILT requires approximately $0.76$ seconds, while CurvyILT incurs a dramatically higher cost of over $4$ seconds.

\vspace{2mm}
\noindent\textbf{Connect optimization-based OPCs with ML models.} We initialize the masks of optimization-based OPC methods using masks generated by different ML baselines. Table~\ref{tab:finetune} summarizes the resulting mask quality and runtime changes. The results show that initialization with MorphOPC consistently achieves the best printing quality, yielding the lowest \textit{$\ell_2$} and \textit{PVB}. In terms of runtime, we observe contrasting behaviors across optimization methods. Initializing MultiILT with model-generated masks leads to a slight slowdown in the optimization process, whereas CurvyILT benefits from such initialization, achieving a modest speedup. Notably, MorphOPC results in the smallest slowdown for MultiILT and the largest speedup for CurvyILT, indicating that it provides the most effective initialization among all compared methods.


\section{conclusion}
In this work, we propose MorphOPC, a geometry-aware OPC model that embeds differentiable morphological operations into the mask generation that align with fundamental lithographic behaviors. Experiments across multiple benchmarks demonstrate that MorphOPC delivers SOTA printability and exhibits strong generalization to unseen layout patterns. 
\bibliographystyle{ACM-Reference-Format}
\bibliography{reference}

\appendix

\end{document}